\documentclass{article}

\usepackage{arxiv}

\usepackage[utf8]{inputenc} 
\usepackage[T1]{fontenc}    
\usepackage{hyperref}       
\usepackage{url}            
\usepackage{booktabs}       
\usepackage{amsfonts}       
\usepackage{nicefrac}       
\usepackage{microtype}      
\usepackage{lipsum}
\usepackage{graphicx}
\usepackage[algo2e]{algorithm2e}
\usepackage{algorithm}
\usepackage{algpseudocode}
\usepackage[countmax]{subfloat}
\usepackage{amsmath}
\usepackage{subcaption}
\usepackage{amssymb} 
\usepackage{pifont}  
\usepackage{multirow}
\usepackage{bbding}
\graphicspath{ {./images/} }

\title{Bi-directional Self-Registration for Misaligned Infrared-Visible Image Fusion}

\author{
 Timing Li \\
  College of Intelligence and Computing\\
  Tianjin University\\
  \texttt{litm@tju.edu.cn} \\
   \And
 Bing Cao \\
  College of Intelligence and Computing\\
  Tianjin University\\
  \texttt{caobing@tju.edu.cn} \\
  \And
 Pengfei Zhu \\
  College of Intelligence and Computing\\
  Tianjin University\\
  \texttt{zhupengfei@tju.edu.cn} \\
  \And
 Bin Xiao \\
  School of Computer Science and Technology\\
  Chongqing University of Posts and Telecommunications\\
  \texttt{xiaobin@cqupt.edu.cn} \\
  \And
 Qinghua Hu \\
  College of Intelligence and Computing\\
  Tianjin University\\
  \texttt{huqinghua@tju.edu.cn} \\
}

\begin{document}
\maketitle
\begin{abstract}
Acquiring accurately aligned multi-modal image pairs is fundamental for achieving high-quality multi-modal image fusion. 
To address the lack of ground truth in current multi-modal image registration and fusion methods, we propose a novel self-supervised \textbf{B}i-directional \textbf{S}elf-\textbf{R}egistration framework (\textbf{B-SR}). Specifically, B-SR utilizes a proxy data generator (PDG) and an inverse proxy data generator (IPDG) to achieve self-supervised global-local registration. Visible-infrared image pairs with spatially misaligned differences are aligned to obtain global differences through the registration module. The same image pairs are processed by PDG such as cropping, flipping, stitching, etc., and then aligned to obtain local differences. IPDG converts the obtained local differences into pseudo-global differences, which are used to perform global-local difference consistency with the global differences. Furthermore, aiming at eliminating the effect of modal gaps on the registration module, we design a neighborhood dynamic alignment loss to achieve cross-modal image edge alignment.
Extensive experiments on misaligned multi-modal images demonstrate the effectiveness of the proposed method in multi-modal image alignment and fusion against the competing methods. Our code will be publicly available.
\end{abstract}


\section{Introduction}
Multi-modal image fusion has become a crucial technology in various fields, including surveillance, medical imaging, and remote sensing \cite{meher2019survey, liu2021learning, yao2019spectral, wang2024rffnet, zhi2018deep}. The infrared images excel at capturing thermal information and identifying objects in low light conditions, while the visible images provide detailed texture and color information. Through the exploitation of inter-modal information complementarity, multi-modal image fusion achieves more accurate and comprehensive scene information \cite{chen2022unsupervised, xuan2022multimodal, zhao2024equivariant, gao2021unified}. 

\begin{figure}[t]
\centering
\includegraphics[width=1\columnwidth]{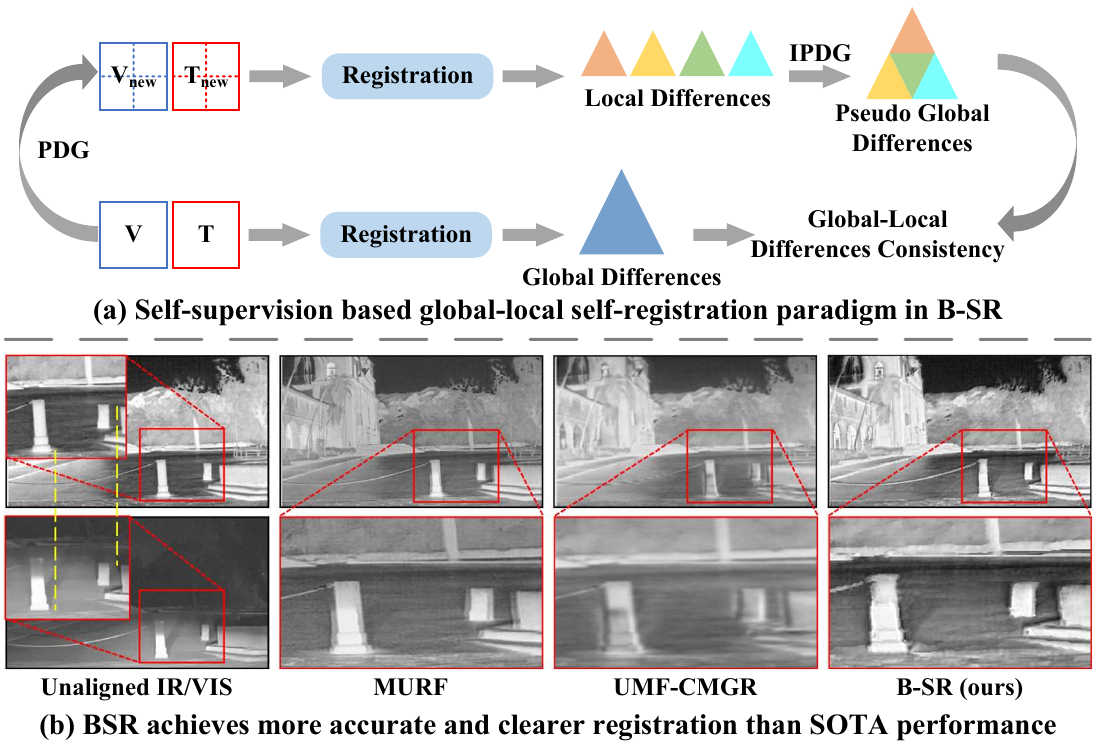}
\caption{(a) Self-registration paradigm in B-SR. (b)
The first column shows infrared and visible image pairs with 5 pixels difference in the horizontal direction, where the yellow dashed line denotes the horizontal difference between different modalities. Compared with existing alignment-based fusion methods, B-SR effectively addresses the issue of misalignment between modalities.
}
\label{fig1}    
\end{figure}

The effectiveness of multi-modal image fusion is heavily based on precise image alignment \cite{li2023feature, tang2022matr, zhou2022promoting, jin2022darkvisionnet, wang2024rffnet, yu2024vifnet}. Spatial misalignment between different modes can lead to degradation of the quality of the fused image, and severe spatial misalignment can result in ghosting in the fused image.\cite{ma2019infrared}. In the acquisition of multi-modal image data, the differences in sensors and camera jitter lead to inevitable changes in the position and viewing angle of different sensors during the image capture process \cite{karim2023current}. This results in spatial differences in the acquired images, and fusion using unaligned multi-modal images can lead to degradation in the quality of the fused image and, in severe cases it can lead to mismatch in the fused image. Unfortunately, current multi-modal fusion methods rarely consider this problem. Therefore, multi-modal image alignment and fusion is an urgent problem to be solved in practical applications.

Most of the existing image fusion methods fail to consider the effect of data misalignment on multi-modal image fusion, 
which leads to significant differences in model performance on different datasets.
As shown in Figure \ref{fig1}, the performance of the model fusion is affected by the quality of the data alignment, i.e., there are significant artifacts in the fusion results after artificially creating non-aligned offsets in the aligned RoadScene dataset. The same problem exists when the model uses the non-aligned DroneVehicle and TNO datasets. In particular, the misalignment problem is especially severe with the drone data due to having a larger viewing angle and more severe jitter problems.

Several related studies have emerged in recent years to mitigate the effects of misaligned data on infrared-visible image fusion. 
These frameworks align multi-modal images using either singer or multiple alignment modules. However, these methodologies invariably rely on image translation, converting one type of multi-modal image into another to reduce the impact of modality discrepancies. These alignment methods based on image translation are straightforward, but inevitably introduce new noise into the images, while making the effectiveness of modal alignment heavily dependent on the performance of image translation. However, the lack of specialized sensors capable of acquiring aligned infrared-visible image pairs represents a significant challenge. Although manual annotation can result in precisely aligned image pairs, the high costs associated with this approach make it difficult to obtain enough aligned data. Therefore, alternative strategies are necessary to supervise the alignment process without ground-truth data.

To address the aforementioned challenges, we propose a novel self-supervised \textbf{B}idirectional \textbf{S}elf-\textbf{R}egistrationtion framework, named \textbf{B-SR}, as shown in Figure \ref{arc}. 
We designed the data proxy generator (PDG) and inverse data proxy generator (IPDG) to achieve global-local self-supervised inter-branch registration, as well as self-supervised intra-branch registration using bi-directional deformation fields. Unlike other methods that rely on image translation models, we propose the neighborhood dynamic alignment loss to reduce the effect of modal differences on the registration module, thus enabling multi-modal image registration without introducing additional noise. Furthermore, the reconstruction module is introduced to bridge the registration and fusion modules, while jointly optimizing the registration and reconstruction modules. Finally, a fused image is generated by fusing the predicted registered image with the original input image.
The main contributions are summarized as follows:

\begin{itemize}
\item We propose a bi-directional self-registration framework that utilizes PDG and IPDG to provide self-supervised signals to the registration network, solving the problem of lacking constraints for the alignment task. 
\item We propose a neighborhood dynamic alignment loss to constrain regions with significant gradient changes, effectively overcoming the modality gap without introducing additional noise typically in translation-based techniques.
\item  
We introduce a self-reconstruction module, jointly optimized with the registration to improve alignment and fusion performance.
Extensive experiments demonstrate our superiority against the competing methods for misaligned infrared and visible image fusion.
\end{itemize}

\section{Related Work}
\label{sec:related_work}

\subsection{Multi-modal Image Alignment}

Image alignment is divided into traditional alignment methods and deep learning alignment methods. Traditional alignment methods primarily include feature-based matching algorithms, mutual information-based registration, and frequency domain alignment techniques \cite{alexey2016discriminative, chen2022sc2}. Arar \cite{arar2020unsupervised} proposed a multi-modal alignment method based on modal translation, which translates different modal images into the same modality before image alignment, reducing the effect of modal differences on multi-modal image alignment. Building upon this foundation, Wang et al. \cite{wang2022unsupervised} proposed an unsupervised alignment network based on modal style transformation, and Han et al. \cite{xu2023murf, xu2022rfnet} further refined this methodology, advancing the research in multi-modal image alignment. While image translation can mitigate modality differences, it inevitably introduces additional random noise \cite{yang2020mri}. Moreover, the dynamic nature of multi-modal images can degrade the performance of image translation. 
Therefore, we propose a dedicated loss function and jointly optimize image alignment and reconstruction tasks without introducing additional noise. This approach aims to achieve image alignment while maximally preserving information from different modalities.


\begin{figure*}[t]
\centering
\includegraphics[width=1\columnwidth]{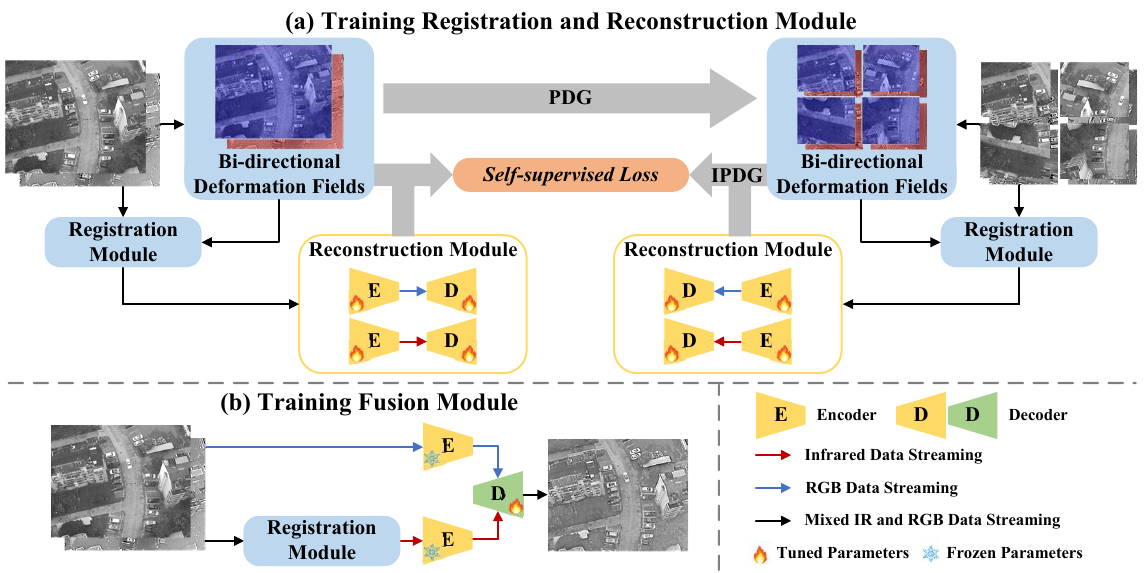}
\caption{Architecture of the B-SR method. (a) The training process of the bi-directional self-registration framework is designed to align multi-modal images by leveraging both intra-branch bi-directional alignment constraints and inter-branch alignment results simultaneously. The inter-branch alignment is achieved through PDG and IPDG, while the intra-branch alignment is achieved through the bi-directional deformation field.  The reconstruction module through joint registration-reconstruction optimization, will fix the feature extraction parameters, thereby providing an efficient feature extraction layer for subsequent fusion. (b) The training process of the fusion module.}
\label{arc}
\end{figure*}

\subsection{Multi-modal Image Fusion}
Based on the powerful feature extraction and reconstruction capabilities, deep learning-based image fusion methods exhibit broad research prospects \cite{ma2020ganmcc, li2021rfn, jung2020unsupervised}. Deep learning-based image fusion approaches can be categorized into generative adversarial networks (GAN), autoencoders (AE), variational autoencoders (VAE), and unified image fusion methods. Ma et al. \cite{ma2019fusiongan} are the first to apply GANs to the image fusion task, designing specialized loss functions to achieve fused images that retain infrared image information while incorporating the texture details of visible light images. Ma et al. \cite{ma2020infrared} designed the detail loss and target edge-enhancement loss to preserve texture details while preserving infrared information. GAN-based image fusion methods can model complex image distributions, thereby enhancing the quality of reconstructed images. Zhao et al. \cite{zhao2020didfuse} decompose multi-modal images into background and detail features before performing feature-level fusion. Zhao et al. \cite{zhao2023cddfuse} proposed an AE-based image fusion method that achieves high-performance image fusion by extracting local and global features. Fabian et al. \cite{duffhauss2022fusionvae} employed a VAE-based approach to generate high-quality fused images by combining multi-modal images with prior knowledge. Unified image fusion-based methods \cite{xu2020u2fusion,zhang2020ifcnn} are usually well generalized and can achieve various image fusion tasks under a unified framework.

\section{Methodology}
\label{sec:Methodology}

We propose a self-supervised bi-directional self-registration framework to address the misalignment issues in multi-modal image fusion, named B-SR. As shown in Fig.~\ref{arc}, the B-SR consists of three components. First, the images of different modalities are aligned with the alignment module, while an image reconstruction module reconstructs both global and local features from disparate-modality images. Subsequently, the self-supervised framework is implemented using proxy data generation (PDG) and inverse proxy data generation (IPDG) to improve the alignment and feature extraction performance. Finally, the image fusion module is used to fuse the aligned multi-modal images. In the following sections, we will provide a detailed explanation of each component.

\subsection{PDG and IPDG}
Lacking ground truth data, we designed two simple and efficient modules, proxy data generator (PDG) and inverse proxy data generator (IPDG), to ensure that B-SR is able to handle different misaligned spatial variances and to increase the diversity of the data. PDG and IPDG are represented as $\mathcal{P}(\cdot)$ and $\mathcal{IP}(\cdot)$, respectively. 

The PDG first crops the image $I$ into $N*N$ patches of the same size. Next, the \emph{i}-th patch (\emph{i} = 1,..., N) is randomly flipped and rotated, and it is important to note that the processing of each patch will be recorded. Finally, the processed patches are re-stitched into a new image $I_{new}$. $I_{new}$ and $I$ will be used as inputs to different branches of B-SR, respectively:
\begin{equation}
\begin{split}
\label{PDG}
I_{new} = \mathcal{P}(I),
\end{split}
\end{equation}

Based on the characteristics of deep learning networks, it is known that the result obtained from inputting $I_{new}$ into the model and then applying an inverse transformation will be the same as the result obtained from inputting $I$ into the model. Therefore, we designed the PDG inverse process IPDG, which is an application to model output operations. Different from the PDG which is only used for images, the IPDG processes the output image $I_{new}$ and the deformation field $\phi_{new}$ between the branches to obtain $I^{'}$ and $\phi^{'}$, respectively, which is formulated as:
\begin{equation}
\begin{split}
\label{IPDG}
I^{'} = \mathcal{IP}(I_{new}), \phi^{'} = \mathcal{IP}(\phi_{new}),
\end{split}
\end{equation}

\subsection{Bi-directional Self-Registration}
To fully exploit the performance of the alignment module, we propose a self-supervised bi-directional self-registration framework. This framework achieves inter-branch alignment and supervision by constraining the results between different branches, while within each branch, a bi-directional alignment structure and a neighborhood dynamic alignment loss are employed to ensure intra-branch alignment. This approach facilitates bi-directional self-registration between infrared and visible images.

Multi-modal image registration is a blind task that lacks accurate ground truth, making it necessary to introduce supervision to improve model registration performance. Therefore, we use PDG to process the misaligned images \emph{T} and \emph{V} to achieve self-supervision model. 

Benefiting from the fact that deep learning has feature symmetry invariance \cite{villar2021scalars, chen2023imaging}, B-SR employs a bi-branch network architecture where image pairs $T$ and $V$, as well as $T_{new}$ and $V_{new}$, are processed separately as inputs. The alignment and reconstruction results of $T_{new}$ and $V_{new}$ are processed through IPDG, implementing a self-supervised registration module, as shown in Fig. \ref{arc}.

In the intra-branch alignment, B-SR employs a U-Net \cite{ronneberger2015u} architecture to generate bi-directional deformation fields $\phi_p$ and $\phi_n$, where $\phi_p$ represents the deformation field from infrared to visible images, and $\phi_n$ denotes the deformation field from visible to infrared images, respectively. To obtain aligned images, we use a spatial transformer network (STN) \cite{jaderberg2015spatial} to deform multi-modal images. The aligned images $\hat{T}$ and $\hat{V}$ are obtained by corresponding deformation field calculations as follows:
\begin{equation}
\begin{split}
\label{stn}
\hat{T} = T \circ \phi_{p}, \hat{V} = V \circ \phi_{n},
\end{split}
\end{equation}
where $\circ$ denotes the spatial transformation. The alignment process for the input values $T_{new}$ and $V_{new}$ from the other branch follows the same principles, resulting in $\hat{T}_{new}$ and $\hat{V}_{new}$. The intra-branch alignment is achieved through imposing constraints on the aligned images within each branch.

Inter-branch alignment, B-SR is achieved by IPDG, which is performed on the alignment results of $T_{new}$ and $V_{new}$, respectively, as shown in the following formula:
\begin{equation}
\begin{split}
\label{IPDG2}
\hat{T}^{'}_{new} = \mathcal{IP}(\hat{T}_{new}), \hat{\phi}^{'} = \mathcal{IP}(\hat{\phi}_{new}),
\end{split}
\end{equation}
Inter-branch alignment is achieved by constraining the inter-branch output images $\hat{T}^{'}_{new}$ and $\hat{T}^{'}$, and the inter-branch deformation fields $\hat{\phi}^{'}_{new}$ and $\hat{\phi}^{'}$, respectively.

\subsection{Image Reconstruction and Fusion Module}
B-SR employs a multi-modal fusion architecture based on an auto-encoder framework \cite{zhao2023cddfuse}. The training process is divided into two stages. In the first stage, we add a decoder for unimodal image reconstruction, allowing the model to reconstruct the registration results in a unimodal manner. This facilitates more comprehensive feature extraction and fusion. In the second training phase, the encoder parameters in the reconstruction module are frozen and the decoder is trained to achieve image fusion.

\subsection{Loss Function.}

\textbf{Neighbourhood dynamic alignment loss.} B-SR differs from existing methods with image translation networks. Therefore, we specifically design a neighborhood dynamic alignment loss, denoted as $\mathcal{L}_{nda}$, for maintaining the consistency of the edge structure of the aligned image with the target image. The $\mathcal{L}_{nda}$  dynamically selects the deformation distance and direction within pixel neighborhoods across different modalities, as shown in Eq. \ref{loss_da}. 
\begin{equation}
\begin{split}
\label{loss_da}
\mathcal{L}_{nda} = \mathcal{L}_{di} + \mathcal{L}_{an},
\end{split}
\end{equation}
where $\mathcal{L}_{di}$ and $\mathcal{L}_{an}$ denote the distance loss of image edges and the angle loss of image edges respectively.
\begin{equation}
\begin{split}
\label{loss_di_an}
\mathcal{L}_{di} = \frac{1}{M}\sum \limits_{i=1}^{M} {\Vert p_{i}(E_e(T), E_e(V)) - p_{i}(E_e(T), E_e(T_a)) \Vert _2},\\
\mathcal{L}_{an} = \frac{1}{M}\sum \limits_{i=1}^{M} {\Vert q_{i}(E_e(T), E_e(V)) - q_{i}(E_e(T), E_e(T_a)) \Vert _1},
\end{split}
\end{equation}
where the image edge $E$ is computed using the Sobel operator, setting $\mu$ as the effective edge threshold, and when $E > \mu$ as the effective edge $E_{e}$. Calculate the effective edges of input images $V$, $T$, and the registered image $T_{a}$, while minimizing both the pixel deformation distance difference $\mathcal{L}_{di}$ as well as the pixel deformation angle difference $\mathcal{L}_{an}$ between these images. $p(\ast)$ and $q(\ast)$ denote the computed distance and angle difference for $\ast$, respectively, and $M$ is the number of effective edges. It is important to clarify that, given the dynamic differences between multi-modal images, $\mathcal{L}_{nda}$ only searches for valid edges of the other modality within the pixel neighborhood. $\mathcal{L}_{nda}$ will constrain the edges in the current modality to align with those of the other modality only when valid edges of the other modality exist within the neighborhood.

\noindent\textbf{Edge pixel retention loss.} To ensure consistency between the pixel information of aligned images and the pixel-wise differences across modalities before alignment, we designed a loss function $\mathcal{L}_{epr}$ to constrain bi-directional self-alignment, as shown in Eq. \ref{loss_lb}.
\begin{equation}
\begin{split}
\label{loss_lb}
\mathcal{L}_{epr} = (T_a - V)^2 - (T - V)^2,
\end{split}
\end{equation}

\noindent\textbf{Self-supervised loss.} Within the self-supervised framework, we employ the $\mathcal{L}_1$ and $\mathcal{L}_2$ as loss functions to constrain the images of the registration module and reconstruction module of the inter-branch,  as shown in Eq. \ref{equ_sem}:
\begin{equation}
\begin{split}
\label{equ_sem}
\mathcal{L}_{ss} = w_1*(\mathcal{L}_1^A(I,I_a) + \mathcal{L}_1^R(I_a,I_r)) +\\
w_2*(\mathcal{L}_2^A(I,I_a) + \mathcal{L}_2^R(I_a,I_r)),
\end{split}
\end{equation}
where $\mathcal{L}_1$ and $\mathcal{L}_2$ are MAE losses and MSE losses, respectively. $\mathcal{L}^A_*$ and $\mathcal{L}^R_*$ denote the constraints on the results of the registration module and the reconstruction module, respectively. $I$ represents infrared and visible images, while $I_a$ denotes the aligned image and $I_r$ denotes the reconstructed image. $w_1$ and $w_2$ denote the weights of the different loss functions.

\noindent\textbf{Smoothing loss.}To enhance the accuracy and smoothness of the deformation field $\phi$ provided by the U-Net architecture, we impose constraints on the spatial gradients of the deformation field $\nabla f(\phi)$.
\begin{equation}
\begin{split}
\label{loss_smooth1}
\mathcal{L}_{smooth} = \sum_{\phi \in \Phi} \Vert \nabla f(\phi) \Vert ^2,
\end{split}
\end{equation}
\begin{equation}
\begin{split}
\label{loss_smooth2}
\nabla f(\phi) = (\frac{\partial f(\phi)}{\partial x} + \frac{\partial f(\phi)}{\partial y}),
\end{split}
\end{equation}
$\Phi \subset \mathbb{R}^2$ is the infrared images and the visible images in the 2-D spatial domain.
Inspired by \cite{balakrishnan2019voxelmorph}, the spatial gradients of displacement $\partial f(\phi)$ is approximated as $\frac{\partial f(\phi)}{\partial x} \approx f((\phi_x + 1,\phi_y)) - f((\phi_x,\phi_y))$ and $\frac{\partial f(\phi)}{\partial y}$ is computed using a similar approximation.


The reconstruction loss is calculated for the reconstructed infrared and visible images to avoid information loss in the reconstructed images. 
\begin{equation}
\begin{split}
\label{loss_recp}
\mathcal{L}_{recp}(I_a) = 1 - \mathcal{L}_{ssim}(I_a - I_r) + \Vert I_a - I_r \Vert _2,
\end{split}
\end{equation}
where $\mathcal{L}_{ssim}$ is the structural similarity index between the image $I_a$ and the reconstructed image $I_r$.

The $\mathcal{L}_{dec}$ constrains the global and local information extraction network. This loss function encourages the preservation of shared global information across different modalities within the same scene while simultaneously promoting the retention of modality-specific local details.
\begin{equation}
\begin{split}
\label{loss_recf}
\mathcal{L}_{recf} = \frac{(cc(F^g_T, (F^g_V))^2}{cc(F^l_T, (F^l_V)+\epsilon},
\end{split}
\end{equation}
where $cc(\ast)$ denotes the correlation coefficient operator, $F^g_{\ast}$ represents the global feature, $F^l_{\ast}$ represents the local feature, and $\epsilon$ is a constant greater than one.


\noindent\textbf{Fusion loss.} Inspired by \cite{tang2022image} and \cite{zhao2023cddfuse}, the loss of the image fusion module is shown in Eq. \ref{loss_fusion}.
\begin{equation}
\begin{split}
\label{loss_fusion}
\mathcal{L}_{fusion} = \mathcal{L}_{dec} + \frac{1}{H*W} \Vert I_F - max(I_T, I_V)\Vert _1 + \\
\frac{1}{H*W} \Vert \vert \nabla I_F \vert - max(\vert \nabla I_T \vert, \vert \nabla I_V \vert) \Vert _1,
\end{split}
\end{equation}
where $H$ and $W$ denote the size of the images and $\nabla$ denotes the Sobel gradient operator.

\section{Experiments}
\label{sec:experiments}

\begin{figure*}[t]
\centering
\includegraphics[width=1\columnwidth]{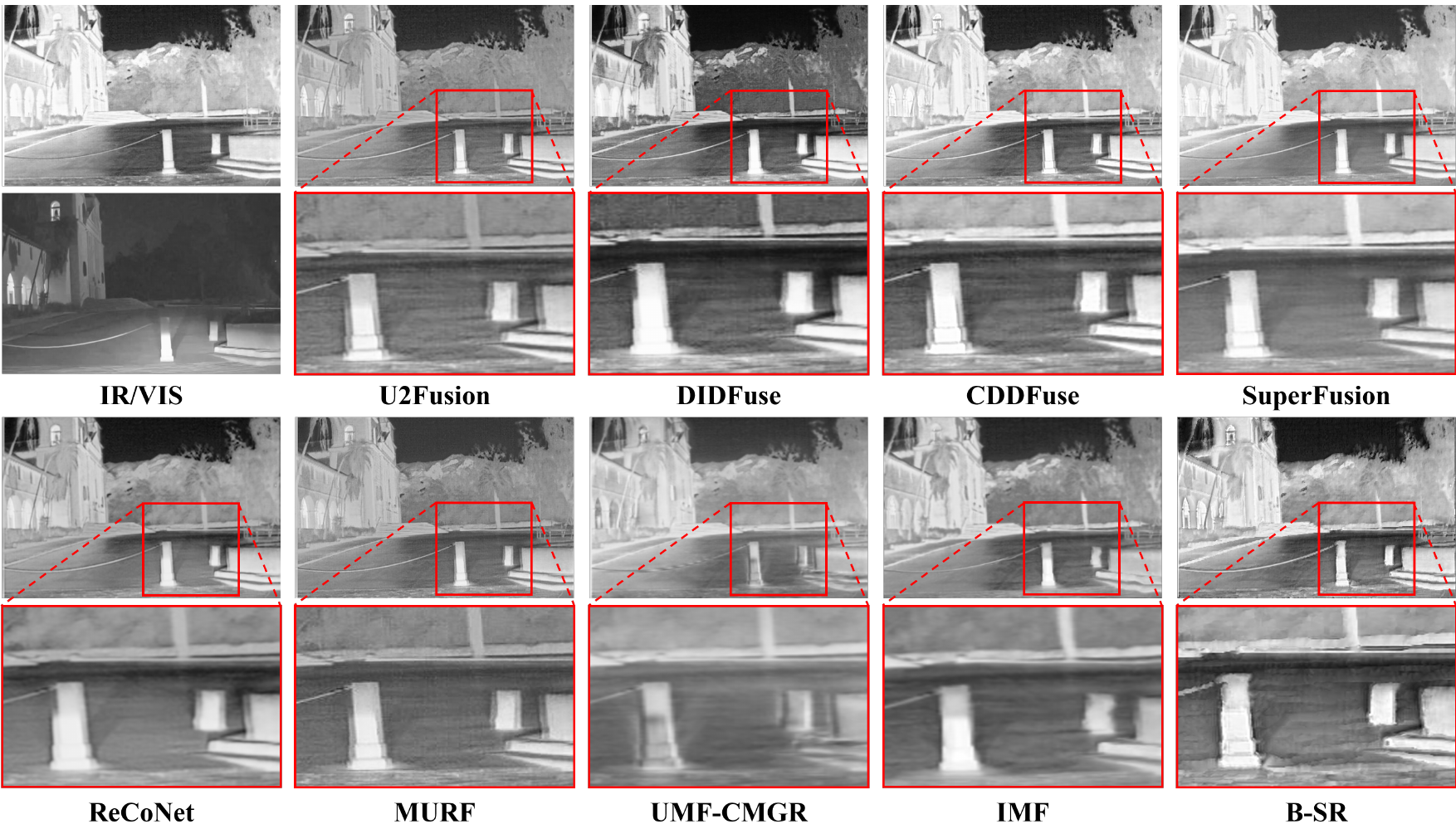} 
\caption{Comparisons of fusion results with a 5-pixel difference in the horizontal direction for "$FLIR\_06993$" in the RoadScene dataset.}
\label{figrs}
\end{figure*}

\subsection{Setup}
\textbf{Datasets.}
We use the popular RoadScene \cite{xu2020aaai} and TNO \cite{toet2012progress} benchmarks to evaluate the performance of our model. Furthermore, to validate the effectiveness of different approaches in handling more complex misaligned multi-modal image fusion scenarios, we conduct experiments on the DroneVehicle \cite{sun2022drone}, which is based on drone views. Given that the RoadScene dataset has been manually aligned, it was necessary to construct misaligned images for both training and testing. we pair visible images with infrared images that have been horizontally shifted by 5 pixels to simulate the spatial discrepancies caused by differences in sensor position.

\noindent \textbf{Metrics.}
We use six metrics to quantitatively measure the alignment and fusion results of the model:  $Q_{abf}$, visual information fidelity(VIFF), spatial frequency (SF), average gradient (AG), mean gradient (MG), and edge intensity (EI). 
These metrics provide the comprehensive evaluation of aligned and fused image quality, detail retention, information integrity, and visual perception performance.

\noindent \textbf{Implement details.}
B-SR needs to be trained for a total of 140 epochs, of which 60 epochs are trained in the first stage and 80 epochs in the second stage. All network parameters are updated with the AdamW optimizer \cite{loshchilov2017fixing} with the initial learning rate set to $10^{-4}$. The parameter $\emph{N}$ in the PDG is 2. The weight parameters $w_1$ and $w_2$ in Eq. \ref{equ_sem} and Eq. \ref{loss_di_an} are 5 and 1, respectively. The effective edge threshold $\mu$ is $10^{-4}$.

\subsection{Comparison with SOTA Methods}
In this section, we evaluate B-SR on three test sets, RoadScene, DroneVehicle, and TNO. We compare the fusion results with state-of-the-art methods including U2Fusion \cite{xu2020u2fusion}, DIDFuse \cite{zhao2020didfuse}, CDDFuse \cite{zhao2023cddfuse}, ReCoNet \cite{huang2022reconet}, SuperFusion \cite{TANG2022SuperFusion}, MURF \cite{xu2023murf}, UMF-CMGR \cite{wang2022unsupervised} and IMF \cite{Wang_2023_IMF}. It is worth noting that, except for U2Fusion, DIDFuse and CDDFuse, the remaining five methods are specifically designed registration modules for the misalignment problem existing in infrared and visible image fusion.

\begin{figure*}[t]
\centering
\includegraphics[width=1\columnwidth]{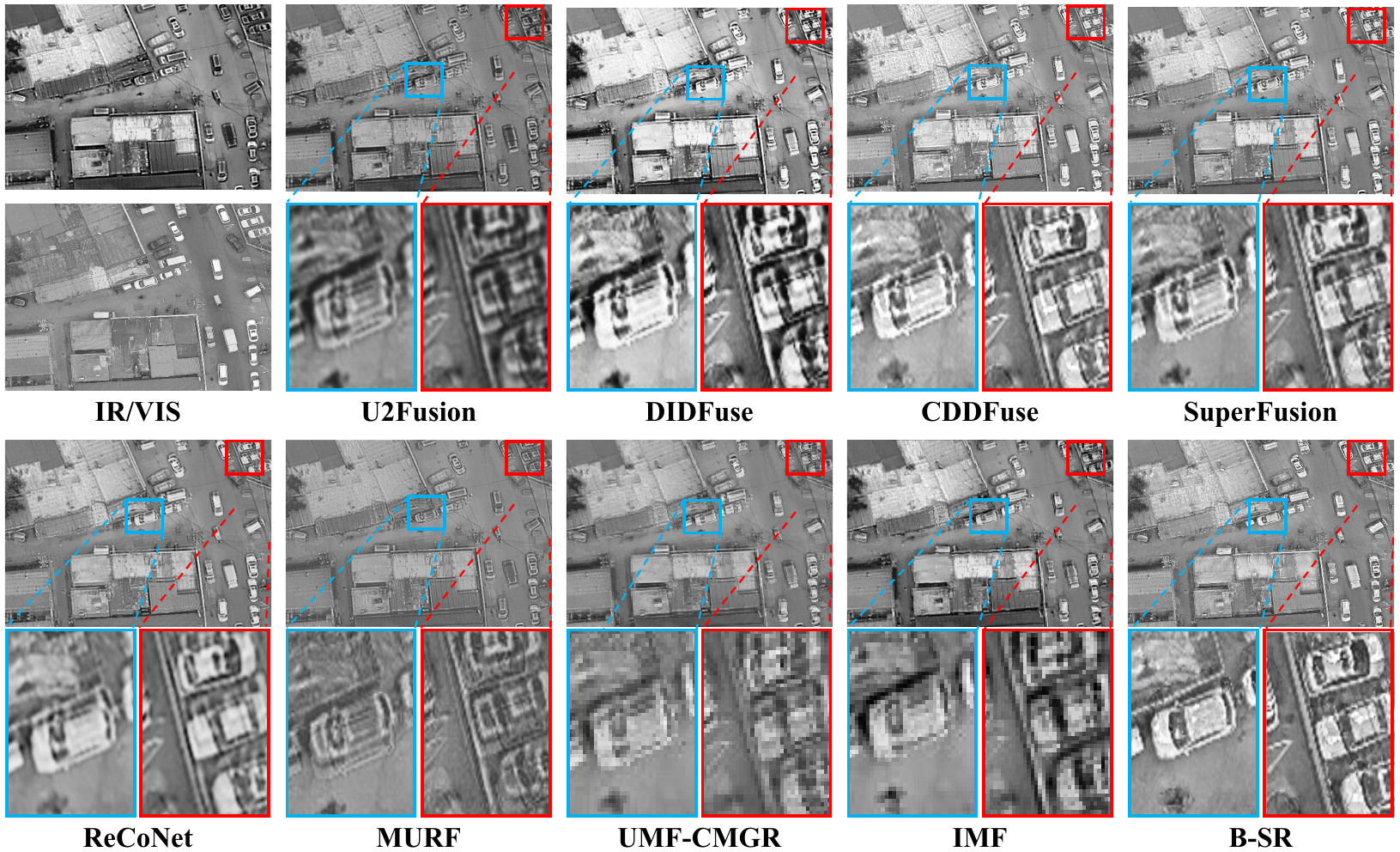} 
\caption{Comparison of results in a DroneVehicle dataset based on drone views.}
\label{figdv}
\end{figure*}

\noindent \textbf{Qualitative comparison.}
We demonstrate qualitative comparisons in different datasets, as shown in Figures \ref{figrs}, \ref{figdv} and \ref{figtno}. In RoadScene \cite{xu2020aaai}, which simulates a 5-pixel difference in the horizontal direction, B-SR can preserve the rich texture information of the image while achieving registration. In the DroneVehicle \cite{sun2022drone} dataset, the large spatial difference leads to obvious ghosting in all the existing registration and fusion methods, but B-SR is able to maintain a good alignment effect. In the TNO \cite{toet2012progress} dataset, where the spatial difference is less, B-SR achieves better results in terms of alignment and texture details.

\noindent \textbf{Quantitative comparison.}
As shown in Tables \ref{table_irleft5}, \ref{table_dv}, and \ref{table_tno}, the results of alignment and fusion are quantitatively assessed using six metrics. In all three datasets, B-SR obtained the highest scores in all six metrics. This shows that B-SR performs well in both the simulated RoadScene dataset with horizontal differences and the misaligned dataset in real scenarios. Meanwhile, B-SR has good alignment and fusion performance in both the DroneVehicle dataset with large spatial differences and the TNO dataset with smaller spatial differences.


\begin{table}[t]
\centering
\caption{Quantitative results of RoadScene with 5 pixel spatial differences in the horizontal direction. Bold indicates the best values.}
\label{table_irleft5}
\resizebox{0.6\columnwidth}{!}{
\Large
\begin{tabular}{c|c c c c c c}
\toprule
Method &Qabf &VIFF &SF &AG &MG &EI \\
\midrule
U2Fusion \cite{xu2020u2fusion} &$0.53$ &$0.57$ &$12.27$ &$4.78$ &$35.89$ &$12.70$\\
DIDFuse \cite{zhao2020didfuse} &$0.27$ &$0.31$ &$16.67$ &$6.00$ &$44.62$ &$15.79$ \\
CDDFuse \cite{zhao2023cddfuse} &$0.51$ &$0.68$ &$17.21$ &$6.14$ &$45.35$ &$16.11$ \\
ReCoNet \cite{huang2022reconet} &$0.40$ &$0.54$ &$8.99$ &$3.63$ &$27.53$ &$14.63$ \\
SuperFusion \cite{TANG2022SuperFusion} &$0.52$ &$0.64$ &$13.08$ &$4.33$ &$32.33$ &$11.45$ \\
MURF \cite{xu2023murf} &$0.49$ &$0.54$ &$13.57$ &$4.69$ &$32.19$ &$11.69$ \\
UMF-CMGR \cite{wang2022unsupervised} &$0.46$ &$0.57$ &$8.77$ &$3.45$ &$25.90$ &$9.18$ \\
IMF \cite{Wang_2023_IMF} &$0.42$ &$0.48$ &$12.78$ &$4.74$ &$35.04$ &$12.50$ \\
\midrule
B-SR &$\mathbf{0.59}$ &$\mathbf{0.72}$ &$\mathbf{18.22}$ &$\mathbf{6.55}$ &$\mathbf{47.74}$ &$\mathbf{17.11}$ \\
\bottomrule
\end{tabular}
}
\end{table}

\begin{table}[t]
\centering
\caption{Quantitative results on the DroneVehicle dataset. Bold indicates the best values.}
\label{table_dv}
\resizebox{0.6\columnwidth}{!}{
\Large
\begin{tabular}{c|c c c c c c}
\toprule
Method &Qabf &VIFF &SF &AG &MG &EI \\
\midrule
U2Fusion \cite{xu2020u2fusion} &$0.28$ &$0.40$ &$10.38$ &$4.19$ &$31.89$ &$13.74$ \\
DIDFuse \cite{zhao2020didfuse} &$0.36$ &$0.50$ &$21.86$ &$7.44$ &$52.33$ &$18.51$ \\
CDDFuse \cite{zhao2023cddfuse} &$0.52$ &$0.62$ &$21.09$ &$7.36$ &$50.87$ &$18.02$ \\
ReCoNet \cite{huang2022reconet} &$0.34$ &$0.43$ &$13.17$ &$5.19$ &$38.95$ &$13.85$ \\
SuperFusion \cite{TANG2022SuperFusion} &$0.53$ &$\mathbf{0.63}$ &$18.40$ &$6.53$ &$46.55$ &$16.48$ \\
MURF \cite{xu2023murf} &$0.37$ &$0.42$ &$14.37$ &$5.41$ &$37.12$ &$13.28$ \\
UMF-CMGR \cite{wang2022unsupervised} &$0.40$ &$0.43$ &$18.71$ &$6.21$ &$43.54$ &$15.46$ \\
IMF \cite{Wang_2023_IMF} &$0.44$ &$0.45$ &$21.75$ &$7.32$ &$51.52$ &$18.36$ \\
\midrule
B-SR &$\mathbf{0.56}$ &$\mathbf{0.63}$ &$\mathbf{24.47}$ &$\mathbf{8.15}$ &$\mathbf{52.96}$ &$\mathbf{18.86}$ \\
\bottomrule
\end{tabular}
}
\end{table}


\subsection{Ablation Studies}
The ablation experiments are performed on misaligned RoadScene to validate the rationality of the different modules. The results of the experiment are shown in Table \ref{table_abe}.

\begin{figure}[t]
\centering
\includegraphics[width=0.6\columnwidth]{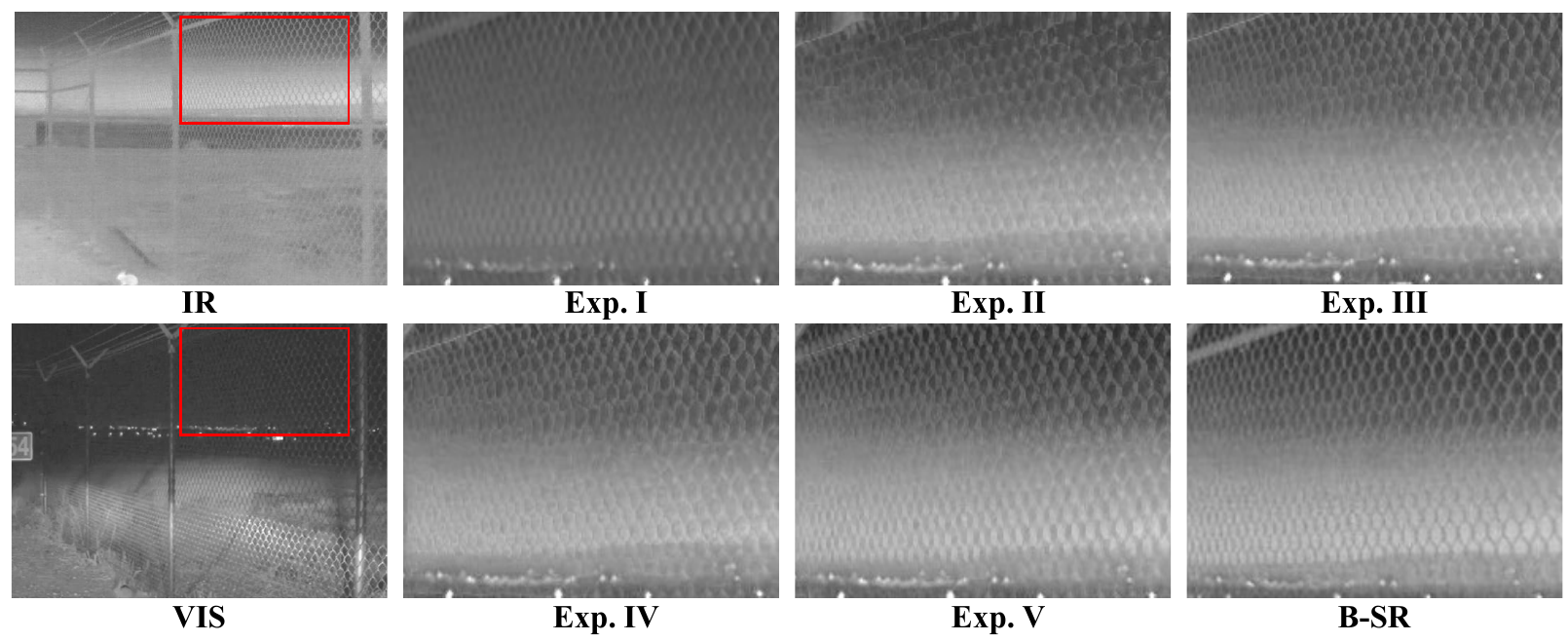} 
\caption{The analysis of the ablation experiment was conducted using the RoadScene dataset.}
\label{figabe}
\end{figure}

\begin{figure*}[t]
\centering
\includegraphics[width=1\columnwidth]{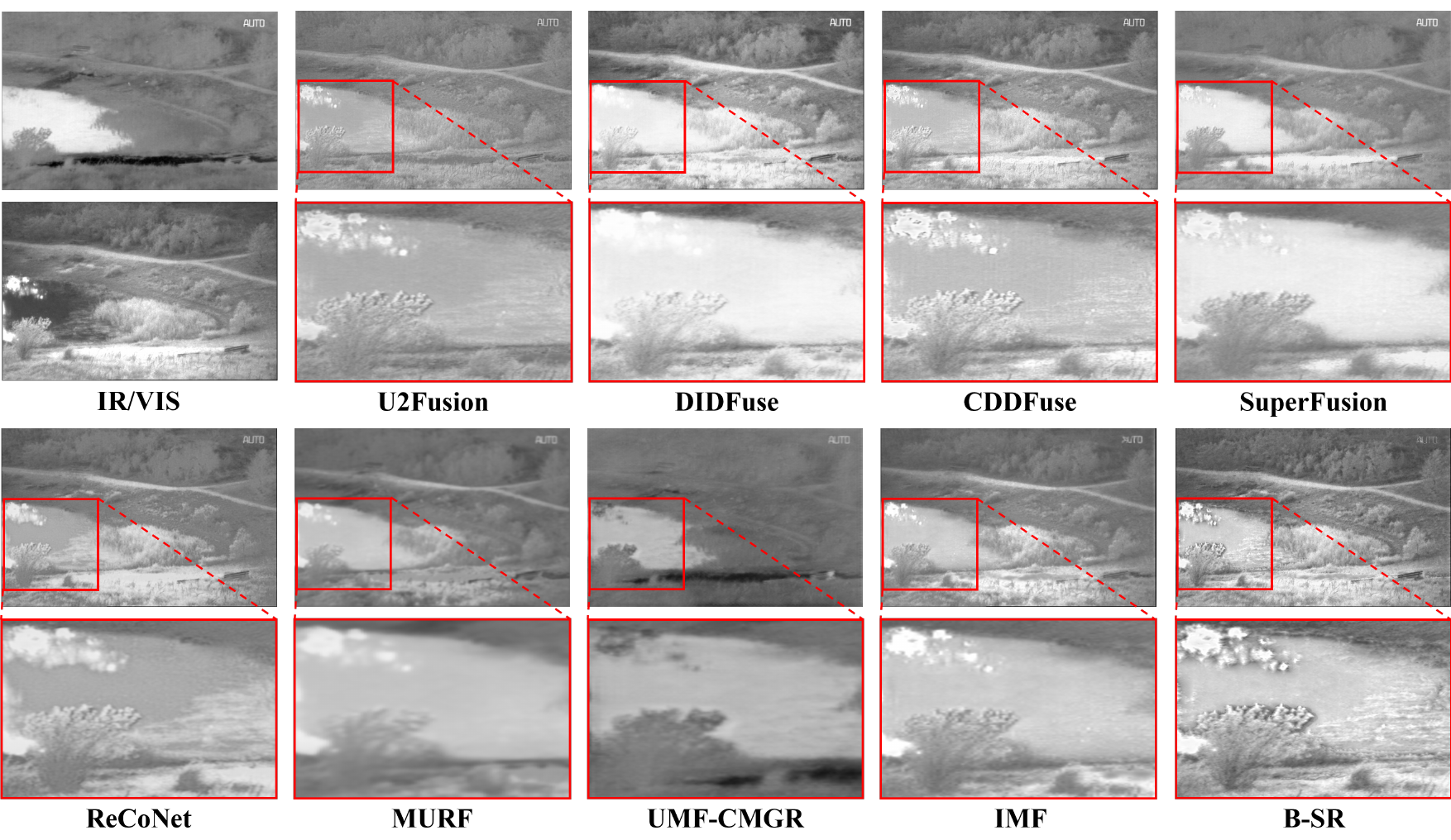} 
\caption{Comparison of fusion results on the TNO dataset.}
\label{figtno}
\end{figure*}

\begin{table}[th]
\centering
\caption{Quantitative results on the TNO dataset.}
\label{table_tno}
\resizebox{0.6\columnwidth}{!}{
\Large
\begin{tabular}{c|c c c c c c}
\toprule
Method &Qabf &VIFF &SF &AG &MG &EI \\
\midrule
U2Fusion \cite{xu2020u2fusion} &$0.45$ &$0.56$ &$10.28$ &$4.20$ &$30.28$ &$10.74$ \\
DIDFuse \cite{zhao2020didfuse} &$0.41$ &$0.61$ &$12.92$ &$4.53$ &$32.07$ &$11.39$ \\
CDDFuse \cite{zhao2023cddfuse} &$0.52$ &$0.74$ &$13.50$ &$5.05$ &$35.01$ &$12.47$ \\
ReCoNet \cite{huang2022reconet} &$0.37$ &$0.53$ &$7.96$ &$3.35$ &$25.01$ &$8.95$ \\
SuperFusion \cite{TANG2022SuperFusion} &$0.45$ &$0.64$ &$9.02$ &$3.37$ &$23.77$ &$8.44$ \\
MURF \cite{xu2023murf} &$0.35$ &$0.42$ &$12.72$ &$4.90$ &$33.99$ &$12.26$ \\
UMF-CMGR \cite{wang2022unsupervised} &$0.43$ &$0.59$ &$3.48$ &$1.47$ &$11.39$ &$4.03$ \\
IMF \cite{Wang_2023_IMF} &$0.46$ &$0.56$ &$14.46$ &$5.07$ &$33.96$ &$12.40$ \\
\midrule
B-SR &$\mathbf{0.56}$ &$\mathbf{0.75}$ &$\mathbf{15.34}$ &$\mathbf{5.50}$ &$\mathbf{37.15}$ &$\mathbf{13.33}$ \\
\bottomrule
\end{tabular}
}
\end{table}

\noindent \textbf{Neighborhood dynamic alignment loss.}
To validate the effectiveness of $\mathcal{L}_{nda}$, we designed Exp. \uppercase\expandafter{\romannumeral1} and Exp. \uppercase\expandafter{\romannumeral2} for separate evaluations. In Exp. \uppercase\expandafter{\romannumeral1}, we replace $\mathcal{L}_{nda}$ with the edge loss of the Sobel operator, while in Exp. \uppercase\expandafter{\romannumeral2}, we removed $\mathcal{L}_{nda}$. Experimental results demonstrate that the absence of the $\mathcal{L}_{nda}$, constraint fails to effectively align image edges across different modalities. In addition, compared to the edge loss of the Sobel operator used, the proposed $\mathcal{L}_{nda}$ offers better alignment performance.

\noindent \textbf{Registration-reconstruction joint optimization (JO).}
In order to illustrate that the joint optimization of the registration module and the reconstruction module can achieve mutual enhancement, in Exp. \uppercase\expandafter{\romannumeral3}, we excluded the reconstruction module from the alignment task. The experimental results show that joint optimization effectively align image edges while preserving image information.

\begin{table}[t]
\centering
\caption{Ablation experiment results in the test set of RoadScene. Bold indicates the best values.}
\label{table_abe}
\resizebox{0.6\columnwidth}{!}{
\Large
\begin{tabular}{c|c c c c c|c c c c}
\toprule
Methods &$\mathcal{L}_{s}$ &$\mathcal{L}_{nda}$ & JO &$\mathcal{L}_{epr}$ &SS & SD & AG & MG & EI \\
\midrule
Exp. \uppercase\expandafter{\romannumeral1} &\ding{51} & &\ding{51} &\ding{51} &\ding{51} &$42.80$  &$5.92$ &$41.67$ &$15.01$ \\
Exp. \uppercase\expandafter{\romannumeral2} & & &\ding{51} &\ding{51} &\ding{51} &$29.65$ &$4.03$  &$29.12$ &$10.33$ \\
Exp. \uppercase\expandafter{\romannumeral3} & &\ding{51} & &\ding{51} &\ding{51} &$42.91$ &$5.02$  &$35.72$ &$12.86$ \\
Exp. \uppercase\expandafter{\romannumeral4} & &\ding{51} &\ding{51} & &\ding{51} &$43.04$ &$5.74$  &$40.79$ &$14.67$ \\
Exp. \uppercase\expandafter{\romannumeral5} & &\ding{51} &\ding{51} &\ding{51} & &$46.52$ &$6.06$  &$42.49$ &$15.54$ \\
\midrule
B-SR & &\ding{51} &\ding{51} &\ding{51} &\ding{51} &$\mathbf{50.42}$ &$\mathbf{6.55}$ &$\mathbf{47.74}$ &$\mathbf{17.11}$ \\
\bottomrule
\end{tabular}
}
\end{table}

\noindent \textbf{Intra-branch bi-directional self-registration.}
In Exp. \uppercase\expandafter{\romannumeral4}, we removed $\phi_n$ and imposed constraints only on the single flow $\phi_p$, without applying the $\mathcal{L}_{epr}$ constraint between $\phi_p$ and $\phi_n$. The experimental results demonstrate that the use of bi-directional self-registration constraints within the alignment network is advantageous for improving the alignment performance of the model.

\noindent \textbf{Self-supervised registration in inter-branching (SS).}
Finally, to demonstrate the effectiveness of branch alignment, we designed Exp. \uppercase\expandafter{\romannumeral5} to evaluate the effect of branch registration employed in the self-supervised framework. The self-supervised framework was removed, and only $\mathcal{L}_{nda}$ and bi-directional self-registration were utilized for intra-branch registration.

\begin{figure}[t]
\centering
\includegraphics[width=0.6\columnwidth]{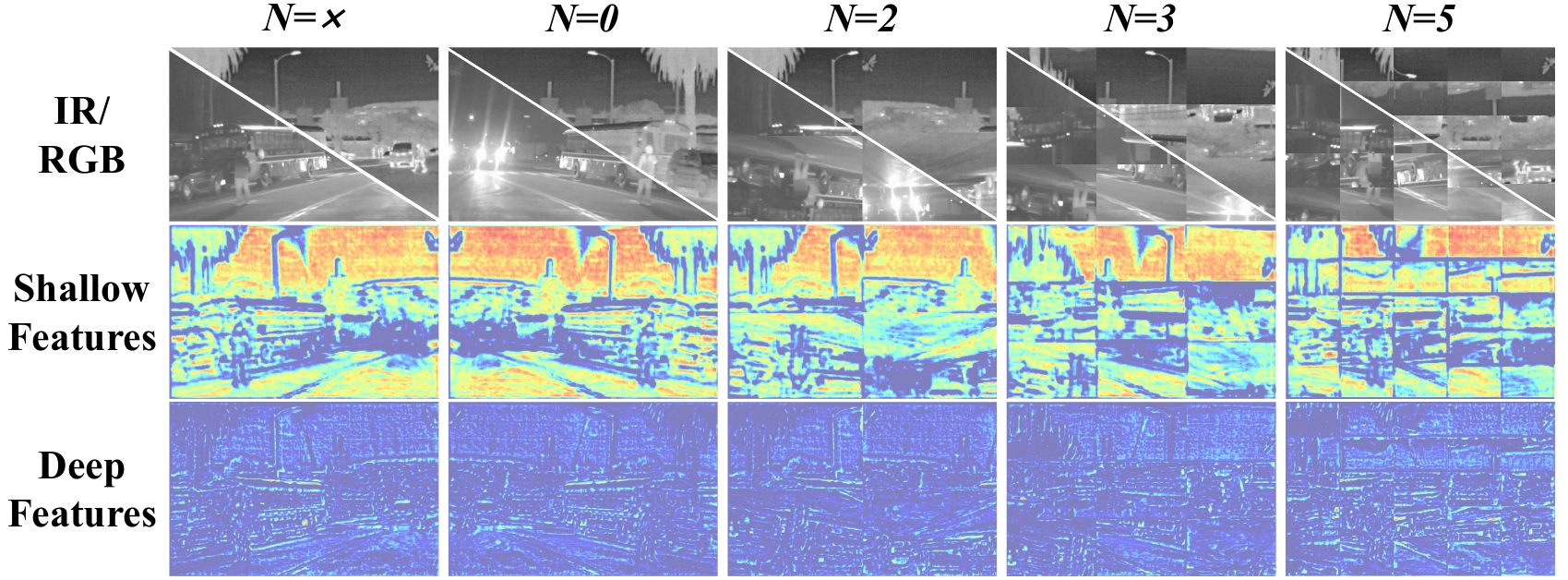} 
\caption{Visualisation of shallow features and deep features after PDG with different parameters $\emph{N}$.}
\label{figheat}
\end{figure}

\subsection{Additional Analysis}
To verify that PDG can effectively assist the model in achieving the registration task, we visualize the features from the registration between the PDG-processed images and the original images. Figure \ref{figheat} illustrates the features of PDG under different parameter $N$ values, where $N = \times$ represents the original input image, and $N = 0$ indicates only random flipping of the original image. When $N$ exceeds 1, the original image is divided into $N*N$ patches followed by random rotation and flipping operations. We dilated the pixel values of infrared images from the RoadScene dataset by 5-pixel and center-cropping.
The feature visualization demonstrates that when $N$ exceeds $1$, the model exhibits enhanced attention to local features rather than global characteristics, thus achieving more precise alignment in localized regions. Simultaneously, the model is compelled to learn more robust feature representations and alignment strategies, facilitating the establishment of potential cross-modal alignment relationships in alignment tasks, thus enhancing the robustness of the self-supervised network. However, as $N$ increases, excessive transformations can introduce pseudo-features, potentially disrupting the local semantic structure and continuity, leading to loss of semantic information. Therefore, $N=2$ represents an optimal balance, maintaining global consistency while improving local alignment capabilities.

\section{Conclusion}
\label{sec:conclusion}

In this paper, we propose a bi-directional self-registration framework to align and fuse infrared and visible image pairs with spatial discrepancies. Our framework solves the problem of the lack of supervised signals for the alignment task by the proxy data generator and the inverse proxy data generator, and achieves self-supervised global-local registration based on self-registration. 
Bi-directional registration is achieved while reducing modal differences by introducing neighborhood dynamic alignment loss and bi-directional deformation fields. This approach fully exploits the performance of the registration network without relying on image translation models. 
By jointly optimizing registration and reconstruction, it achieves the alignment of infrared-visible image pairs while maintaining image integrity. 
Experimental results demonstrate that our B-SR method improves the registration and fusion performance regardless of whether the data exhibits are obviously or slightly misaligned. 

\bibliographystyle{unsrt}  
\bibliography{references}  






\clearpage
\setcounter{page}{1}

In the supplementary material, we provide more details, more experimental results, and more analysis of ``Bi-directional Self-Registration for Misaligned Infrared-Visible Image Fusion''. 
\begin{itemize}
\item We discuss the differences between our proposed bidirectional self-registration framework and existing alignment-based image fusion methods.
\item We describe the network structure as well as the neighborhood dynamic alignment loss in detail.
\item To demonstrate our method copes with various types of misaligned infrared-visible images, we have conducted additional image experiments on dilated images for more complex misalignment scenarios. 
\item To demonstrate the robustness of our method, we also perform experiments for different levels of dilation and center cropping misalignment. 
\item We have analyzed statistical distributions of feature maps with different parameters of the proxy data generator (PDG). 
\end{itemize}

\section{More Details}

\subsection{Pseudocode of B-SR}

We provide implementation details for each of the two training phases of B-SR to clearly describe the training process of B-SR, as shown in Algorithm \ref{pc1}.

\begin{algorithm}[t]

\caption{Pseudocode for B-SR training phase.}
\KwIn{\rm origin \ multi-modal \ images \ \it V, \ T \\}
\KwOut{\rm fusion \ images \it F \\ }

\BlankLine
\textbf{Phase 1: Image Registration and Reconstruction}\\
\For{\ {V, T} \ \textbf{in} \ Dataloader \ }{
$V_{new}, T_{new} = \mathcal{P}(V), \mathcal{P}(T)$ \hfill // Acquisition of self-supervised data via PDG

$\phi_{p}, \phi_{n} = UNet(V,T)$ \hfill // Generate bi-directional deformation fields

$\phi_{p-new}, \phi_{n-new} = UNet(V_{new},T_{new})$ \hfill // Generate bi-directional deformation fields

$\hat{V} = V \circ \phi_{n}, \hat{T} = T \circ \phi_{p} $ \hfill // Generate aligned images

$\hat{V}_{new} = V_{new} \circ \phi_{n-new}, \hat{T}_{new} = T_{new} \circ \phi_{p-new} $ \hfill // Generate aligned images

$\hat{V}^{'}_{new}, \hat{T}^{'}_{new} = \mathcal{IP}(\hat{T}_{new}), \mathcal{IP}(\hat{V}_{new})$ \hfill // Generate pseudo-global images via IPDG

$\hat{\phi}^{'}_{p}, \hat{\phi}^{'}_{n} = \mathcal{IP}(\hat{\phi}_{p-new}), \mathcal{IP}(\hat{\phi}_{n-new})$ \hfill // Generate pseudo-global deformation fields via IPDG

$\hat{V}^{R} =\rm Decoder(\rm Encoder(\it \hat{V} \rm)\rm )$\hfill // Generate reconstruction images

$\hat{T}^{R}\ =\rm Decoder(\rm Encoder(\it \hat{T} \rm))$ 

$\hat{V}^{R'}_{new} =\rm Decoder(\rm Encoder(\it \hat{V}^{'}_{new} \rm)\rm )$

$\hat{T}^{R'}_{new}\ = \rm Decoder(\rm Encoder(\it \hat{T}^{'}_{new} \rm)\rm )$

$\mathcal{L}_{nda}(V,T,\hat{V}), \mathcal{L}_{nda}(V_{new},T_{new},\hat{V}_{new})$ \hfill // Calculate neighborhood dynamic loss $\mathcal{L}_{nda}$

$\mathcal{L}_{epr}(V,T,\hat{V}), \mathcal{L}_{epr}(V_{new},T_{new},\hat{V}_{new})$ \hfill // Calculate edge pixel retention loss $\mathcal{L}_{epr}$

$\mathcal{L}_{ss}(\hat{T}^{R}, \hat{T}^{R}, \hat{V}^{R'}_{new}, \hat{T}^{R'}_{new}, \phi_{p-new}, \phi_{n-new}, \hat{\phi}^{'}_{p}, \hat{\phi}^{'}_{n})$ \hfill // Global-local differences consistency

}

\BlankLine
\textbf{Phase 2: Image Fusion}\\
\For{\ {V, T} \ \textbf{in} \ Dataloader \ }{
$\hat{T} = T \circ \phi_{p}$ \hfill // Acquire the aligned infrared image

$F =\rm Decoder(\rm Encoder(\it V \rm) + Encoder(\it \hat{T} \rm)\rm )$\hfill // Freeze Encoder parameters to train Decoder

\label{pc1}
}

\end{algorithm}

\subsection{Framework Comparisons}

\begin{figure}[t]
    \centering
    \begin{subfigure}[t]{1.0\columnwidth}
        \centering
        \includegraphics[width=\textwidth]{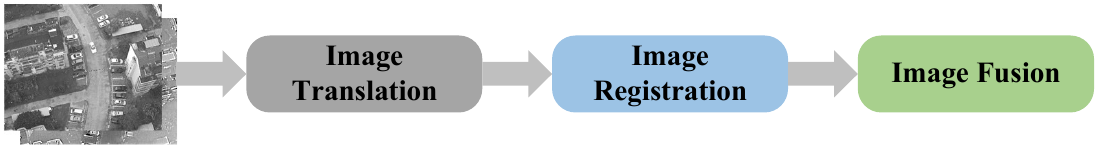}
        \caption{Single alignment framework}
        \label{f2a}
    \end{subfigure}

    \begin{subfigure}[t]{1.0\columnwidth}
        \centering
        \includegraphics[width=\textwidth]{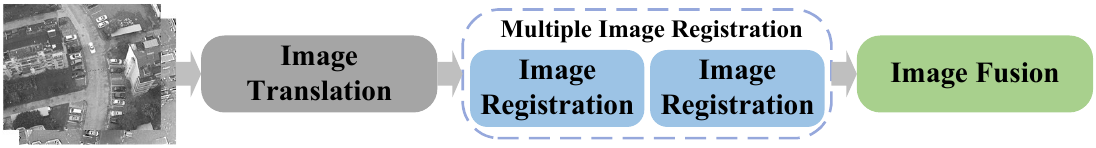}
        \caption{Multiple alignment framework}
        \label{f2b}
    \end{subfigure}

    \begin{subfigure}[t]{1.0\columnwidth}
        \centering
        \includegraphics[width=\textwidth]{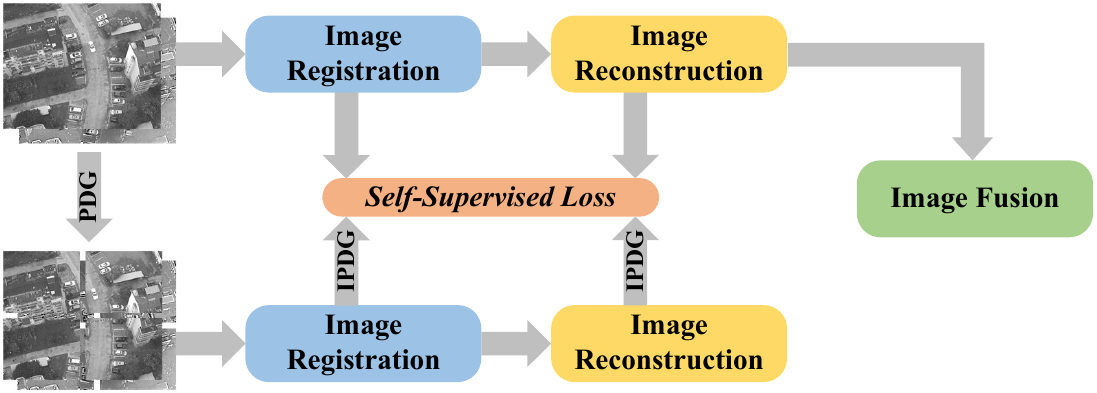}
        \caption{Proposed framework}
        \label{f2c}
    \end{subfigure}

    \caption{The proposed bi-directional self-registration framework is compared with existing single-alignment frameworks and multiple-alignment frameworks.}
    \label{fig2}
\end{figure}

Existing multi-modal fusion methods based on image alignment can be categorized into single-alignment frameworks (SAF) \cite{Wang_2023_IMF, wang2022unsupervised} and multi-alignment frameworks (MAF) \cite{xu2023murf}, as shown in Figure~\ref{f2a} and Figure~\ref{f2b}. The common characteristic of the SAF and MAF approaches is that both require image translation of the original data. However, the inherently dynamic nature of infrared and visible images degrades the performance of image translation when dealing with targets present in the current modality (e.g., infrared images) but lacking a corresponding representation in the target modality (e.g., visible images).

Our proposed bi-directional self-registration framework (B-SR), as shown in Figure \ref{f2c}, differs from existing frameworks in several key aspects. B-SR addresses the lack of ground truth in multi-modal alignment tasks through the proxy data generator (PDG) and inverse proxy data generator (IPDG), while also eliminating the dependency on image translation networks. This design ensures that the registration and fusion performance of the model is not limited by image translation quality. Consequently, it also reduces the risk of introducing additional noise from image translation networks. The gap of multi-modal images can be addressed by the proposed neighborhood dynamic loss.

\begin{figure*}[t]
\centering
\includegraphics[width=1\columnwidth]{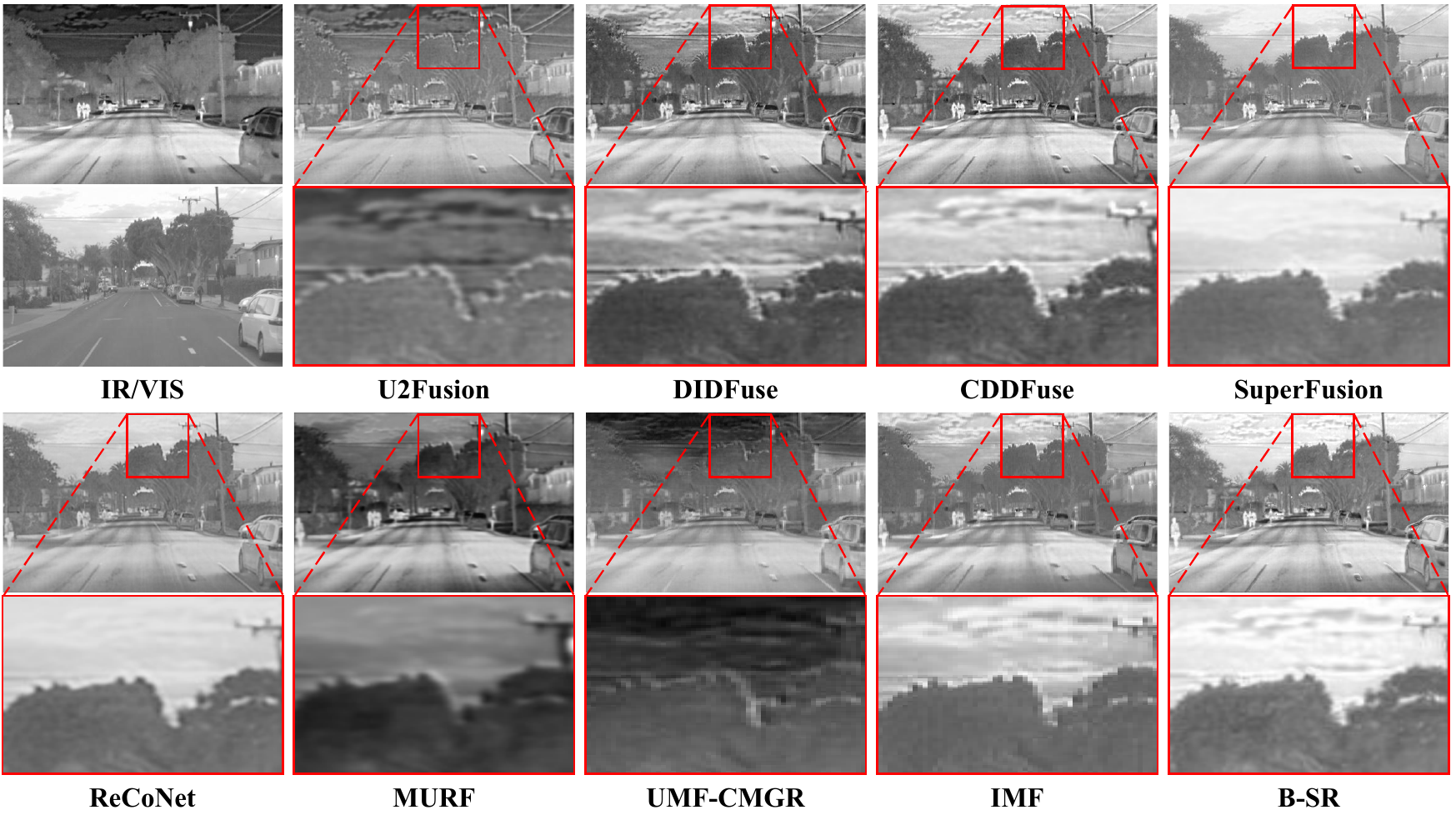} 
\caption{Comparison of RoadScene results with misaligned data after dilation and center cropping.}
\label{figrc}
\end{figure*}

\subsection{More Details of Image Reconstruction and Fusion Module}
Considering that multi-modal image fusion is a blind task lacking ground truth, it is impossible to perform supervised training on the fusion results. To address this issue, B-SR adopts a multi-modal image fusion strategy based on an autoencoder (AE), decomposing the image fusion process into two components: image reconstruction and image fusion. The image reconstruction module results are used to impose supervisory constraints on both the encoder and decoder. After training the image reconstruction module, the parameters of the encoders for different modalities are frozen and used as feature extractors during the fusion phase. In fusion phase, the decoder is continued to be trained, with the input consisting of fused features from the different modalities.

\subsection{More Details of Loss Functions}
\noindent \textbf{Inter-Branch Registration Loss.}
Our B-SR is a bi-directional self-registration framework that achieves a self-supervised structure via PDG and IPDG. The branch without PDG processing obtains global differences. The branch with PDG processing obtains local differences, and pseudo-global differences are obtained by IPDG processing of the local differences. Consider that the results between the two model branches should be equivalent. In other words, the global differences are the same as the pseudo-global differences. Therefore, we achieve global-local discrepancy consistency due to the imposition of $l_1$ and $l_2$ losses on the deformation fields, alignment results, and reconstruction results between branches, respectively.

\noindent \textbf{Intra-Branch Registration Loss.}
The intra-branch registration loss consists of two components, $L_{nda}$ and $L_{epr}$. 

First, the neighborhood dynamic alignment constraint $L_{nda}$ is designed to constrain the edges of images in different modalities. The $L_{nda}$ contains $L_{di}$ and $L_{an}$ constraints from the aligned pixel-level shift distance and angle, respectively. Given that infrared images have less texture information compared to visible images, an effective edge threshold is established to reduce unnecessary computations. The effective edges of different modal images are computed by constraining the distance and direction angle of their effective edges. We calculate the distance-angle difference between the effective boundaries of the infrared and visible images before alignment, as well as the difference between the aligned infrared image and the original infrared image. By constraining the distance and direction of the effective edges in the neighborhood of the pixel points, alignment between the infrared and visible images is achieved when the sum of the distance and direction reaches its minimum.

Second, we design the loss of preservation of the bi-directional layout difference, $L_{epr}$, to maintain the structural difference between the edges of images before and after alignment. B-SR calculates the Euclidean distance $O_1$ between the edges of the pre-alignment infrared image $T_1$ and the visible image. Similarly, it calculates the Euclidean distance $O_2$ between the edges of the post-alignment infrared image $T_2$ and the pre-alignment infrared image $T_1$. By minimizing the difference between $O_1$ and $O_2$, the structural differences in the aligned images are preserved. In particular, it should be noted that different from the edge distance loss $L_{di}$ in $L_{nda}$, $L_{epr}$ focuses on preserving the structural integrity of the images during and after alignment, ensuring that the inherent differences in the layout are maintained even after the images are aligned. On the other hand, $L_{di}$ is about achieving accurate alignment by dynamically constraining the neighborhood around pixel points, ensuring the edges align well between the different modalities.

\section{More Experiments}

\subsection{Setup}
\label{sec:rs_rc}
To ensure that B-SR can adapt to various misalignment scenarios, we simulated a dilation and center cropping misalignment by dilating the infrared images in the aligned RoadScene dataset by 5 pixels using interpolation and then cropped the dilated images based on their center to restore them to their original size. Similar to previous experiments, we trained using the same sequence of image pairs, with each pair being cropped to one-quarter of its original size. We use six metrics to quantitatively measure the alignment and fusion results of the model: $Q_{abf}$, visual information fidelity (VIFF), spatial frequency (SF), average gradient (AG), mean gradient (MG), and edge intensity (EI).

\subsection{Comparison with SOTA methods}
In this section, we simulate new types of misalignment, \emph{i.e.}, experiments using RoadScene datasets that have been processed with dilation and center cropping. We compare the fusion results with state-of-the-art methods including U2Fusion \cite{xu2020u2fusion}, DIDFuse \cite{zhao2020didfuse}, CDDFuse \cite{zhao2023cddfuse}, ReCoNet \cite{huang2022reconet}, SuperFusion \cite{TANG2022SuperFusion}, MURF \cite{xu2023murf}, UMF-CMGR \cite{wang2022unsupervised} and IMF \cite{Wang_2023_IMF}.

\begin{table}[t]
\centering
\caption{Quantitative results for RoadScene with misaligned data after dilation and center cropping. Bold indicates the best values.}
\label{table_rc30}
\resizebox{0.6\columnwidth}{!}{
\Large
\begin{tabular}{c|c c c c c c}
\toprule
Method &Qabf &VIFF &SF &AG &MG &EI \\
\midrule
U2Fusion \cite{xu2020u2fusion} &$0.45$ &$0.56$ &$9.17$ &$3.92$ &$30.28$ &$10.71$ \\
DIDFuse \cite{zhao2020didfuse} &$0.28$ &$0.34$ &$16.57$ &$6.04$ &$45.12$ &$15.97$ \\
CDDFuse \cite{zhao2023cddfuse} &$0.57$ &$0.71$ &$15.13$ &$5.85$ &$43.63$ &$15.50$ \\
ReCoNet \cite{huang2022reconet} &$0.41$ &$0.56$ &$8.30$ &$3.38$ &$25.67$ &$9.15$ \\
SuperFusion \cite{TANG2022SuperFusion} &$0.50$ &$0.64$ &$10.67$ &$3.86$ &$28.97$ &$10.26$ \\
MURF \cite{xu2023murf} &$0.55$ &$0.62$ &$7.93$ &$3.33$ &$25.20$ &$9.09$ \\
UMF-CMGR \cite{wang2022unsupervised} &$0.26$ &$0.42$ &$5.76$ &$2.54$ &$19.81$ &$7.02$ \\
IMF \cite{Wang_2023_IMF} &$0.43$ &$0.48$ &$13.01$ &$4.69$ &$34.98$ &$12.42$ \\
\midrule
B-SR &$\mathbf{0.59}$ &$\mathbf{0.72}$ &$\mathbf{18.44}$ &$\mathbf{6.51}$ &$\mathbf{47.42}$ &$\mathbf{16.94}$ \\
\bottomrule
\end{tabular}
}
\end{table}

\noindent \textbf{Qualitative Comparison.}
In the qualitative results, B-SR demonstrates the ability to align image edges while preserving image information, as shown in Figure \ref{figrc}. Compared to alignment-based fusion methods, B-SR achieves superior alignment and produces clearer fused images. It significantly improves edge alignment while maintaining fusion quality on par with or better than other fusion methods.

\noindent \textbf{Quantitative Comparison.}
As shown in Table \ref{table_rc30}, the results of alignment and fusion are quantitatively assessed using six metrics. In the context of dilation and center cropping misaligned datasets, B-SR obtained the highest scores in all six metrics. The results show that B-SR still performs well in the face of dilation and center cropping misaligned.



\begin{figure*}[t]
\centering
\includegraphics[width=1\columnwidth]{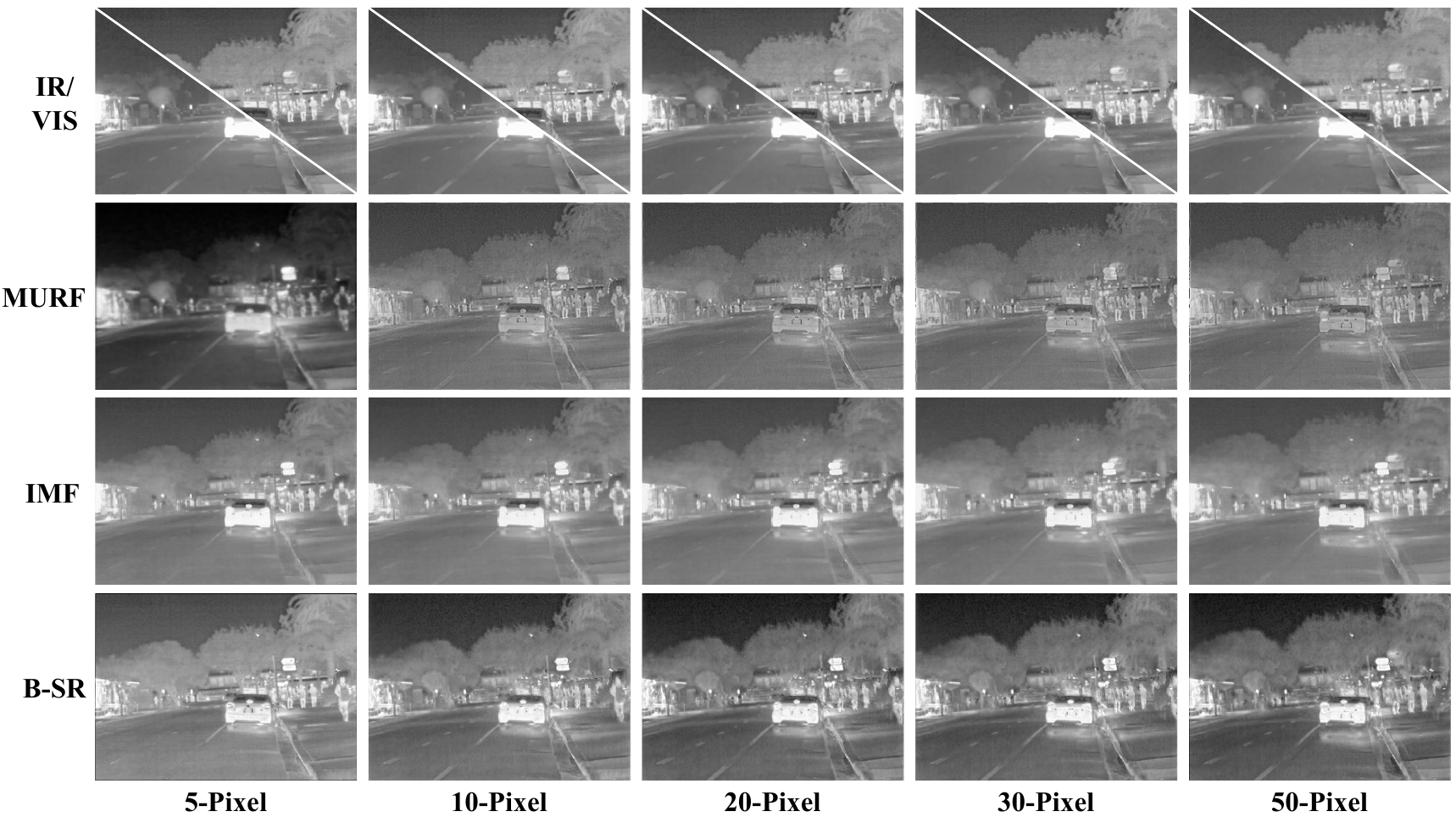} 
\caption{ The results of registration and fusion under unaligned data with different levels of dilation and center cropping.}
\label{rc_pixel}
\end{figure*}

\subsection{Robustness Analysis}
To validate the robustness of B-SR, we evaluated the model trained on the dataset described in Section \ref{sec:rs_rc} using test data with varying degrees of dilation and center cropping misalignment. The magnitudes of dilation and center cropping deformation were set to 5, 10, 20, 30, and 50 pixels, respectively. The experimental results are shown in Figure \ref{rc_pixel}, where the B-SR still preserves the edge registration when the infrared image is deformed by 50 pixels of dilation and center cropping deformation. In quantitative tests conducted under different levels of misalignment, B-SR still maintains good performance metrics, indicating that the B-SR has good robustness.

\begin{table}[t]
\centering
\caption{Quantitative results for RoadScene with misaligned data after dilation and center cropping.}
\label{table_rc_pixel}
\resizebox{0.6\columnwidth}{!}{
\large
\begin{tabular}{c|c|c c c c c c}
\toprule
Level &Method &Qabf &VIFF &SF &AG &MG &EI \\
\midrule
\multirow{3}{*}{5-pixel} 
 &MURF \cite{xu2023murf} &$0.55$ &$0.62$ &$7.93$ &$3.33$ &$25.20$ &$9.09$ \\
 &IMF \cite{Wang_2023_IMF} &$0.43$ &$0.48$ &$13.01$ &$4.69$ &$34.98$ &$12.42$ \\
 &BSR &$0.59$ &$0.72$ &$18.44$ &$6.51$ &$47.42$ &$16.94$ \\
\midrule
\multirow{3}{*}{10-pixel} 
 &MURF \cite{xu2023murf} &$0.47$ &$0.50$ &$14.39$ &$5.41$ &$38.75$ &$13.08$ \\
 &IMF \cite{Wang_2023_IMF} &$0.40$ &$0.45$ &$12.71$ &$4.64$ &$34.54$ &$12.28$ \\
 &BSR &$0.59$ &$0.76$ &$15.79$ &$5.47$ &$39.16$ &$14.00$ \\
\midrule
\multirow{3}{*}{20-pixel} 
 &MURF \cite{xu2023murf} &$0.47$ &$0.51$ &$14.21$ &$5.49$ &$38.73$ &$14.07$ \\
 &IMF \cite{Wang_2023_IMF} &$0.41$ &$0.45$ &$12.73$ &$4.68$ &$34.94$ &$12.42$ \\
 &BSR &$0.59$ &$0.78$ &$15.71$ &$5.50$ &$39.49$ &$14.12$ \\
\midrule
\multirow{3}{*}{30-pixel} 
 &MURF \cite{xu2023murf} &$0.47$ &$0.52$ &$13.97$ &$5.42$ &$38.32$ &$13.53$ \\
 &IMF \cite{Wang_2023_IMF} &$0.41$ &$0.46$ &$12.61$ &$4.66$ &$34.84$ &$12.38$ \\
 &BSR &$0.59$ &$0.79$ &$15.44$ &$5.45$ &$39.15$ &$14.00$ \\
\midrule
\multirow{3}{*}{50-pixel} 
 &MURF \cite{xu2023murf} &$0.47$ &$0.52$ &$13.59$ &$5.29$ &$37.46$ &$13.62$ \\
 &IMF \cite{Wang_2023_IMF} &$0.41$ &$0.47$ &$12.40$ &$4.59$ &$34.40$ &$12.22$ \\
 &BSR &$0.60$ &$0.80$ &$15.23$ &$5.42$ &$39.04$ &$13.96$ \\
\bottomrule
\end{tabular}
}
\end{table}

\subsection{More Discussions on PDG}
\begin{figure}[t]
\centering
\includegraphics[width=0.6\columnwidth]{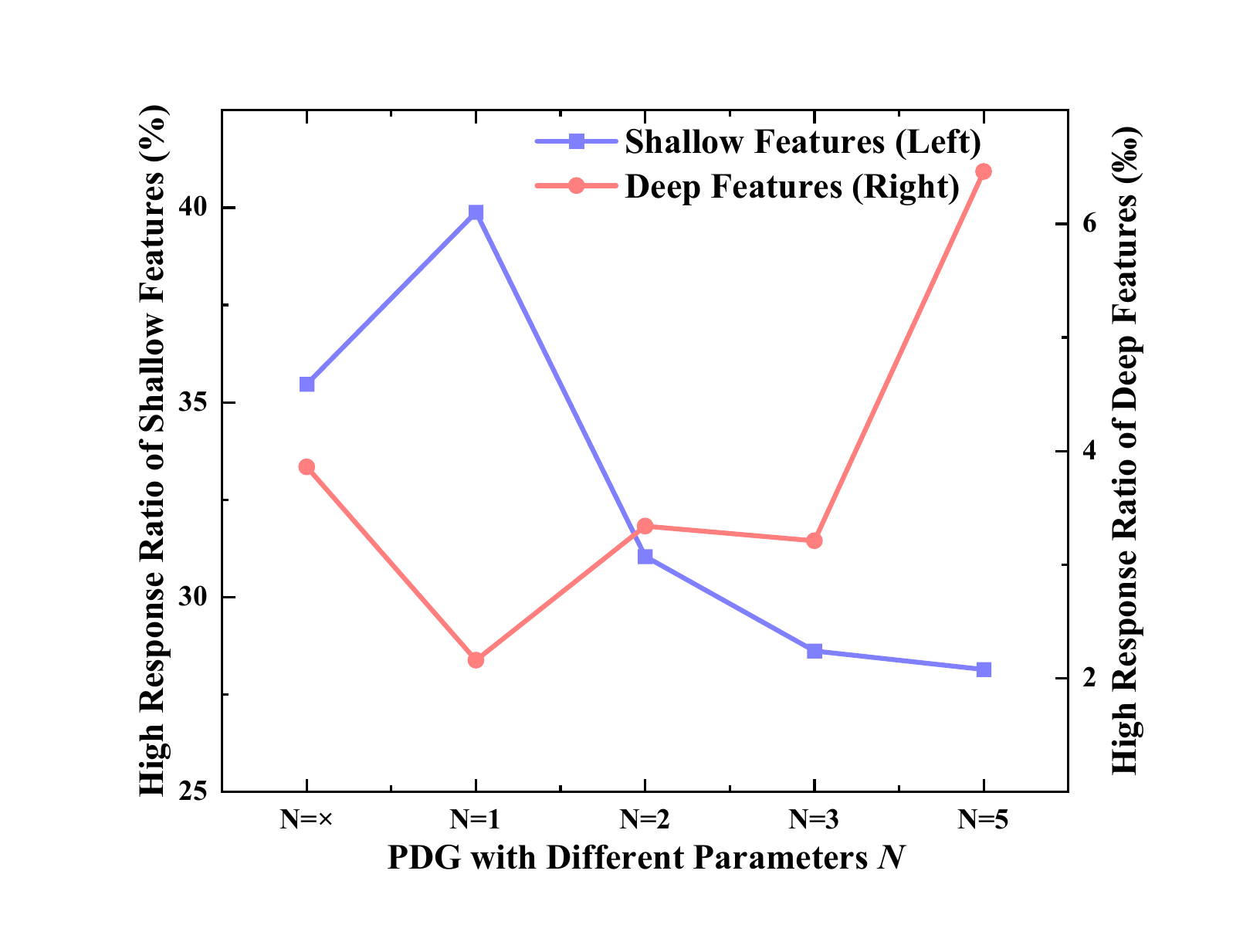} 
\caption{Percentage of high response areas in the features.}
\label{figshm}
\end{figure}

In this section, we statistically compare the highly responsive areas of the parameter $N$ in PDG in deep features and shallow features. The red and orange regions in the feature maps are the highly responsive areas, and we statistically compare the percentage of this area in the whole image. Figure \ref{figshm} shows the percentage of highly responsive areas in PDG for different parameter $N$, where $N = \times$ represents the original input image, and $N = 0$ indicates only random flipping of the original image. When $N$ exceeds 1, the original image is divided into $N*N$ patches followed by random rotation and flipping operations. We dilated the pixel values of infrared images from the RoadScene dataset by 5-pixel and center-cropping.

For shallow features, although the image being cropped can increase the model's focus on local features, the cropping leads to a decrease in the continuity of the image, which lowers the model's response value at shallow features. For deep features, deep features pay more attention to semantic features and registration information of key areas. Cropping more patches benefits the model to integrate semantic information at smaller local information and improves the response value of the model. Therefore, when $N$ is too small, it leads to a high shallow response value of the model, but insufficient local alignment ability and deep alignment feature extraction. On the contrary, when $N$ is too large, it leads to low model shallow response values due to the introduction of too many local features, resulting in a model that lacks global structural comprehension. Taken together, the model strikes a balance between global and local features when $N = 2$.

\end{document}